\documentclass{article}

\usepackage{arxiv}

\usepackage[utf8]{inputenc} 
\usepackage[T1]{fontenc}    
\usepackage{hyperref}       
\usepackage{url}            
\usepackage{booktabs}       
\usepackage{amsfonts}       
\usepackage{nicefrac}       
\usepackage{microtype}      
\usepackage{cleveref}       
\usepackage{lipsum}         
\usepackage{graphicx}
\usepackage{natbib}
\usepackage{doi}

\title{A Biophysically Detailed \textit{C.~elegans} Circuit as a Task-Agnostic Dynamical Core for Visually Robust Robot Manipulation}

\date{}

\newif\ifuniqueAffiliation

\ifuniqueAffiliation 
\author{ \href{https://orcid.org/0000-0000-0000-0000}{\includegraphics[scale=0.06]{orcid.pdf}\hspace{1mm}David S.~Hippocampus}\thanks{Use footnote for providing further
		information about author (webpage, alternative
		address)---\emph{not} for acknowledging funding agencies.} \\
	Department of Computer Science\\
	Cranberry-Lemon University\\
	Pittsburgh, PA 15213 \\
	\texttt{hippo@cs.cranberry-lemon.edu} \\
	\And
	\href{https://orcid.org/0000-0000-0000-0000}{\includegraphics[scale=0.06]{orcid.pdf}\hspace{1mm}Elias D.~Striatum} \\
	Department of Electrical Engineering\\
	Mount-Sheikh University\\
	Santa Narimana, Levand \\
	\texttt{stariate@ee.mount-sheikh.edu} \\
}
\else
\usepackage{authblk}

\makeatletter
\newif\if@cofirst@done
\@cofirst@donefalse
\DeclareRobustCommand{\cofirst}{%
  \footnotemark[2]%
  \if@cofirst@done\else
    \@cofirst@donetrue
    \protected@xdef\@thanks{\@thanks
      \protect\footnotetext[2]{These authors contributed equally.}}%
  \fi
}
\DeclareRobustCommand{\corresponding}[1]{%
  \footnotemark[1]%
  \protected@xdef\@thanks{\@thanks
    \protect\footnotetext[1]{#1}}%
}
\makeatother

\author[1]{%
	Linrui Qian\cofirst%
}
\author[1]{%
	Jiajia Zhang\cofirst%
}
\author[1]{%
	Gan He%
}
\author[1]{%
	Bohan Sun%
}
\author[1,2]{%
	Zhiwei Lin%
}
\author[1]{%
	Qianhao Wang%
}
\author[1]{%
	Zewu Cai%
}
\author[2]{%
	Nianyu Yi%
}
\author[3]{%
    Mengdi Zhao%
}
\author[4]{%
	Kai Du\thanks{Correspondence to: \texttt{kai\_du@tsinghua.edu.cn}}%
}
\affil[1]{CogLeap.AI Space Intelligence (Wuxi) Technology Co., Ltd., Beijing 100080, China}
\affil[2]{School of Mathematics and Computational Science, Xiangtan University, Xiangtan 411105, China}
\affil[3]{Institute for Brain and Intelligence, Fudan University, Shanghai 200433, China.}
\affil[4]{Department of Psychological and Cognitive Sciences, Tsinghua University, Beijing 100084, China}
\fi

\renewcommand{\shorttitle}{\textit{C.~elegans} circuit for visually robust manipulation}

\hypersetup{
pdftitle={A biophysically detailed \textit{C.~elegans} sensorimotor circuit as a task-agnostic dynamical core for visually robust robot manipulation},
pdfsubject={q-bio.NC, q-bio.QM},
pdfauthor={Linrui Qian, Jiajia Zhang, Gan He, Bohan Sun, Zhiwei Lin, Qianhao Wang, Zewu Cai, Kai Du},
pdfkeywords={\textit{Caenorhabditis elegans}, biophysically detailed model, task-agnostic dynamical core, visual robustness, neural manifold},
}

\graphicspath{{figures/}}

\begin{document}
\maketitle

\begin{abstract}
Robot policies are usually trained for one task, one body and one visual environment, and generalize poorly beyond these conditions. Whether a nervous system can instead supply the sensorimotor computation through its evolved wiring and biophysics remains unresolved. Here we embed a biophysically detailed \textit{Caenorhabditis elegans} sensorimotor circuit --- 136 multicompartment neurons with realistic morphologies and electrophysiological characteristics --- as the dynamical core of a visuomotor policy. Only thin task-specific adapters are trained; the core's synaptic weights stay fixed while its membrane voltages evolve freely. Across different MetaWorld tasks the core matches or exceeds diffusion-policy, action-chunking-transformer and neural-circuit-policy baselines, and degrades less under visual perturbations. Replacing the core with generic network models such as MLP, LSTM, transformer or reservoir networks removes the advantage. Furthermore, on a real robotic arm the core withstands diverse visual perturbations that collapse the baselines. Our results suggest that visual robustness can be inherited from biophysically detailed circuit dynamics rather than learned by task-specific controllers.
\end{abstract}



\section{Introduction}

Robot learning has usually converged on a familiar recipe: a task-specific sensory encoder, a high-capacity temporal model, and a task-specific action decoder, all trained together on demonstrations of the task at hand. This recipe is powerful but structurally fragile. Because the sensorimotor computation lives inside parameters that were fitted to one task, one body and one visual environment, every new setting invites retraining, and closed-loop stability under distribution shift is difficult to guarantee a prior. A large literature compensates with data augmentation, domain randomization or explicit system dentification \citep{tobin2017domain,hendrycks2019benchmarking}. An alternative is to inherit the computation itself: to take a nervous system whose connectivity and dynamics were shaped by evolution, reuse it with its parameters held fixed, and thereby obtain the dynamical core of many controllers.

The nematode \textit{Caenorhabditis elegans} is the natural test case. Its 302 neurons and their chemical and electrical synapses have been mapped completely \citep{white1986connectome}, the wiring is non-random and exhibits hierarchical, modular and rich-club organization \citep{varshney2011structural,towlson2013rich}, and whole-brain calcium imaging now provides population-level activity that constrains models of its dynamics \citep{cook2019whole}. BAAIWorm integrates these constraints into a multicompartment brain model with realistic morphology that is closed onto a body--environment simulator \citep{zhao2024baaiworm}. Whether such a circuit, detached from the worm's own body, can act as a task-agnostic dynamical core for an
articulated robot is an open question with consequences for both robotics and neuroscience: it asks whether evolved wiring encodes a computational prior that survives a change of embodiment.

Here we test that hypothesis directly. We reuse the complete 136-neuron sensorimotor circuit of BAAIWorm, with its 1{,}901 inter-neuron connections (1{,}115 chemical synapses and 786 gap junctions) and its
biophysically detailed multicompartment neuron models, as a dynamical core whose synaptic weights are fixed and whose membrane voltages evolve freely in continuous time. RGB images are compressed by a task-specific convolutional encoder into a low-dimensional sensory drive that is injected into the circuit as clamp currents; the resulting motor-neuron state is conditioned by FiLM and decoded into a short action chunk. Only the encoder, FiLM parameters and decoder are trained, and the same fixed-parameter \textit{C.~elegans} core is shared by every task.

We report three findings. First, the arrangement is task-agnostic: across simulated manipulation tasks the core performs competitively and degrades less under visual perturbation than diffusion-policy (DP) \citep{chi2023diffusion}, action-chunking-transformer (ACT) \citep{zhao2023act} and neural-circuit-policy (NCP) \citep{lechner2020ncp} systems. Second, the advantage is attributable to the core rather than to the adapters, its size or generic recurrence: with the interface, data and training protocol held fixed, replacing the core by MLP, LSTM, transformer or reservoir alternatives removes it. Third, the effect of robustness and ablation effects transfer well to a real robot.

Our claims are bounded deliberately. We do not claim zero-shot transfer to unseen tasks, generalisation across arbitrary bodies or cameras, or that the present experiments exhaust the space of perturbations. We also stress that BAAIWorm is a published model that we reuse, not a model we introduce. What we claim is that a biophysically detailed, biologically constrained sensorimotor circuit can act as a task-agnostic dynamical core for robot manipulation, that its contribution is localized to the circuit that generic recurrent cores do not reproduce.

\begin{figure}[htbp]
\centering
\includegraphics[width=\textwidth]{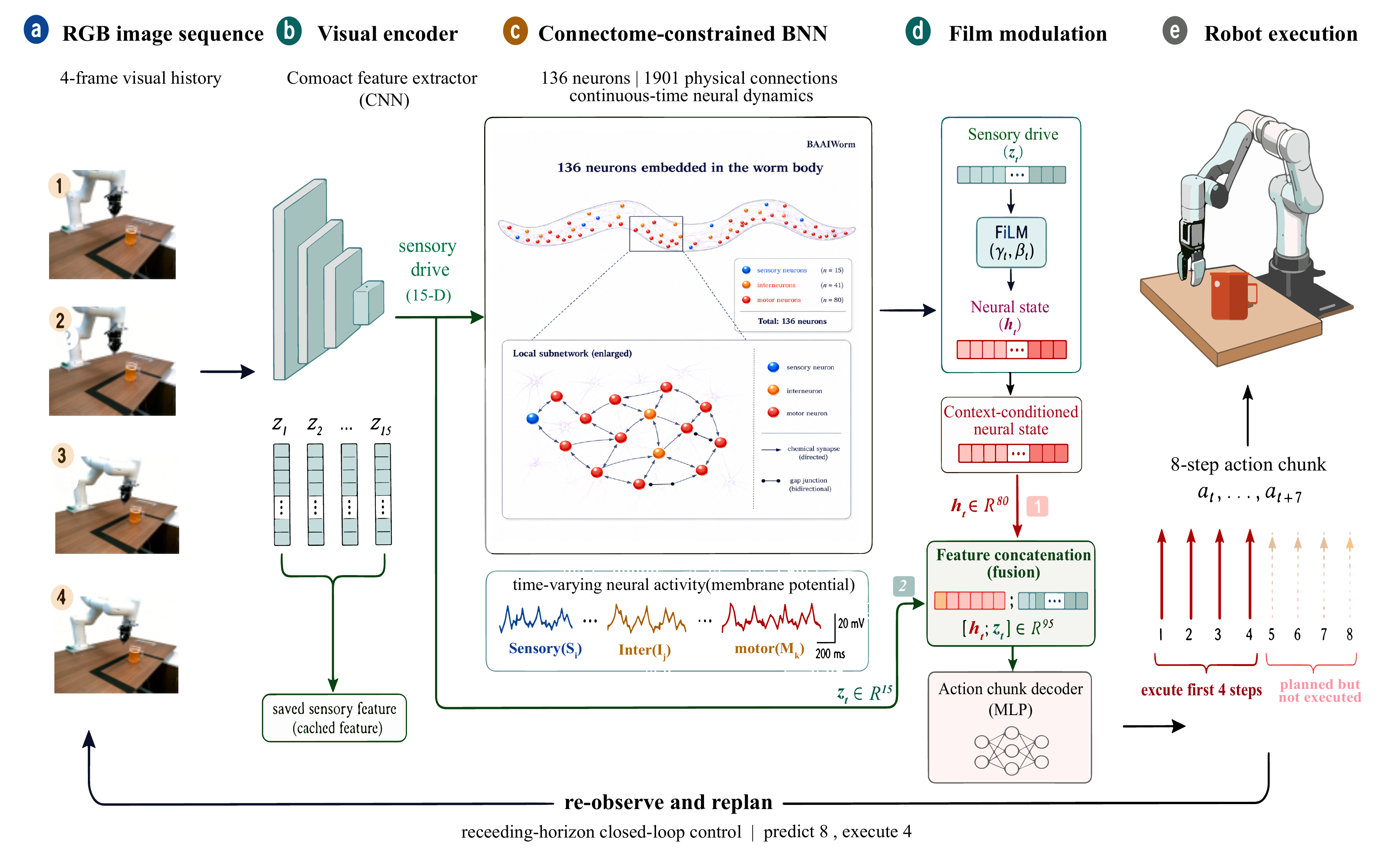}
\caption{\textbf{A biophysically detailed \textit{C.~elegans} sensorimotor circuit as the task-agnostic dynamical core of a visuomotor policy.}
Continuous RGB frames are compressed into a low-dimensional sensory drive that is injected into the connectome-constrained circuit as clamp currents. The circuit's motor-neuron state is conditioned with
visual features by FiLM and decoded into an action chunk that is executed in a receding-horizon loop (predict eight steps, execute four, re-observe and replan). The circuit contains 136 multicompartment
neurons and 1{,}901 interneuron connections; its parameters are fixed after training and its membrane voltages evolve freely, and only the encoder, FiLM parameters and action decoder are trained. The circuit
is the complete sensorimotor model of BAAIWorm \citep{zhao2024baaiworm}, reused here without modification.}
\label{fig1}
\end{figure}


\section{Results}

\subsection{A biophysically detailed sensorimotor circuit as a task-agnostic dynamical core}

The system is defined in figure~\ref{fig1}. A four-frame RGB history is encoded by a compact CNN into a 15-dimensional sensory drive. Rather than being fed to a task network, the drive is converted into clamp currents and injected into the 136-neuron circuit, whose neurons are multicompartment models with realistic dendrites and active conductances, including voltage-gated calcium channels and
calcium-regulated potassium channels. The circuit evolves in continuous time; its 80 motor neurons form a motor neural state that is modulated by FiLM and concatenated with the visual feature, and a small decoder produces an eight-step action chunk of which four steps are executed before replanning. Only the encoder, the FiLM parameters and the decoder are trained, and the same core is used for every task.
The design therefore separates two roles that are usually entangled: a task-specific interface that maps observations to the core's input space, and a task-agnostic dynamical core that generates the
sensorimotor trajectories.

\begin{figure}[htbp]
\centering
\includegraphics[width=0.9\textwidth]{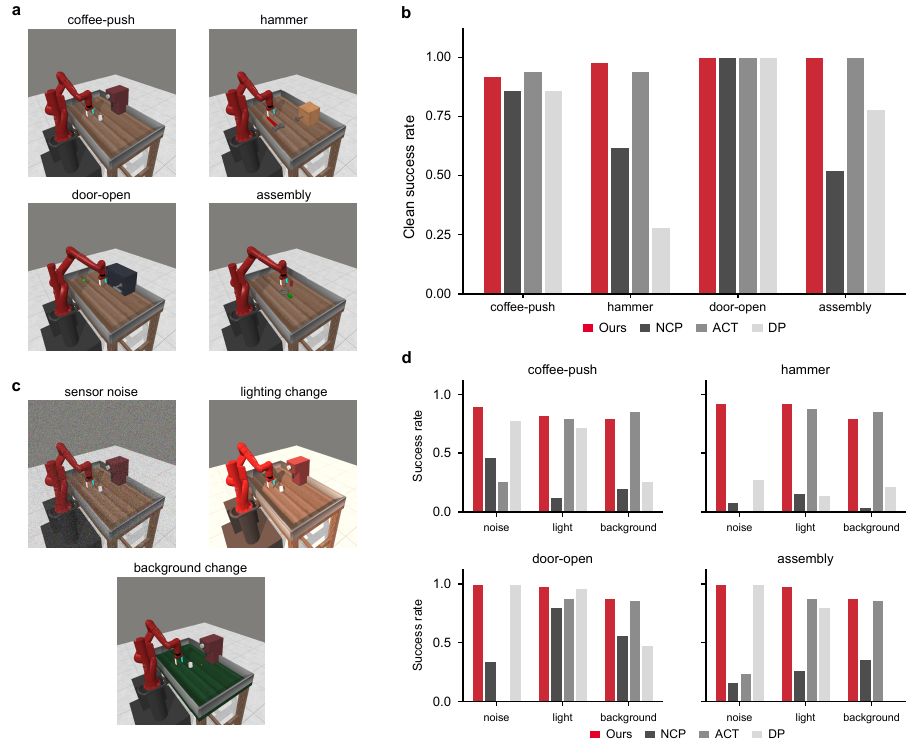}
\caption{\textbf{A single core across four simulated manipulation tasks and three perturbations.}
\textbf{a}, Simulator frames of the four tasks (coffee-push, hammer, door-open, assembly). \textbf{b}, Clean success rate of the core, NCP, ACT and DP; the core matches or exceeds every baseline. \textbf{c}, The perturbations, shown on simulator frames: additive image noise ($\sigma = 0.10$), lighting change, background change and object displacement. \textbf{d}, Success rate under each perturbation, one panel per task; within each tick the bars are Ours, NCP, ACT and DP.}
\label{fig2}
\end{figure}

\subsection{A single core sustains diverse manipulation tasks under visual perturbation}

We first asked whether a single core can be shared by several manipulation tasks. Four MetaWorld tasks were evaluated with the same core and task-specific adapters: coffee-push, hammer, door-open and
assembly. Clean success rates of the core matched or exceeded DP, ACT and NCP in every task (Fig.~\ref{fig2}b): coffee-push 0.92 versus 0.86 (DP), 0.94 (ACT) and 0.86 (NCP); hammer 0.98 versus 0.28, 0.94 and 0.62; door-open 1.00 versus 1.00, 1.00 and 1.00; assembly 1.00 versus 0.78, 1.00 and 0.52.

Under perturbation the differences became qualitative rather than incremental (Fig.~\ref{fig2}c,d). With pixel-level Gaussian noise (single intensity, $\sigma = 0.10$) the core was almost unchanged (0.90--1.00 across the four tasks), whereas ACT, NCP and DP fell to 0.00--0.26, 0.08--0.46 and 0.28--1.00. Under illumination change the core retained 0.82--0.98 while ACT, DP and NCP dropped to 0.80--0.88, 0.14--0.96 and 0.12--0.80, and under background change the core (0.80--0.88) and ACT (0.86) stayed largely intact while DP (0.00--0.48) and NCP (0.04--0.56) collapsed. The core's advantage is therefore not confined to a single task or perturbation type, and it is not explained by a better clean policy: on three of four tasks
the clean rates were already at ceiling for every method.

\subsection{Robustness is carried by the circuit, not by the interface}

\begin{figure}[htbp]
\centering
\includegraphics[width=\textwidth]{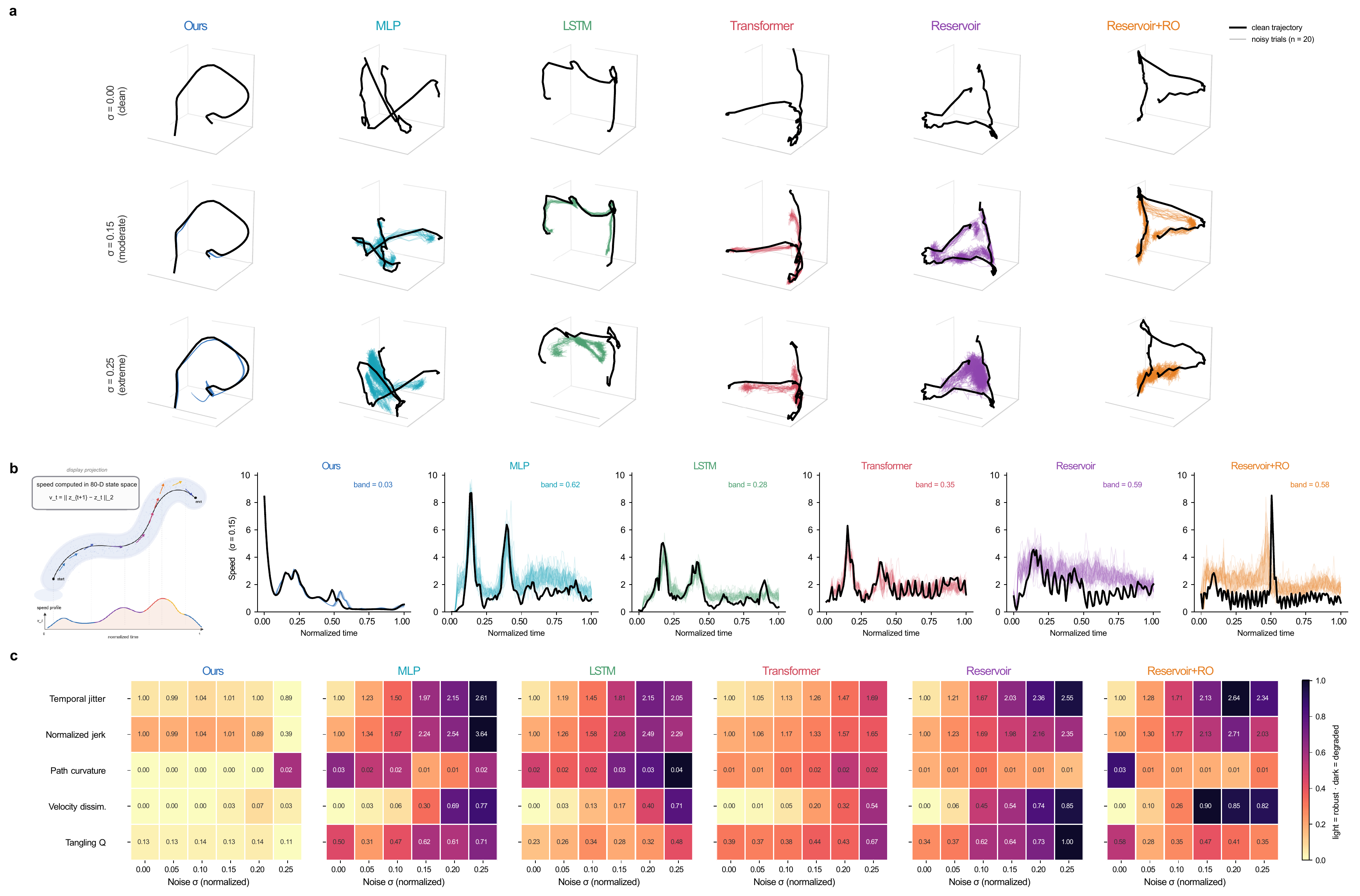}
\caption{\textbf{Robustness is carried by the circuit, not by the interface.}
\textbf{a}, Motor-state trajectories of the full circuit and five replacement cores at three noise levels, projected onto a manifold fitted to the clean trajectories of each model. Clean trajectories are black; colored lines are individual noisy trials. \textbf{b}, Speed profiles on the manifold for moderate noise ($\sigma = 0.15$), with the band width between the noisy and clean profiles reported above each panel. \textbf{c}, Five dynamical metrics (temporal jitter, normalized jerk, path curvature, velocity dissimilarity and tangling) as a function of the noise level, normalized so that 1 marks the clean
reference. All variants share the same encoder, FiLM mechanism, decoder and action representation; only the core differs. }
\label{fig3}
\end{figure}

Because the adapter is trained per task, a natural objection is that the observed robustness comes from the interface rather than from the circuit. Figure~\ref{fig3} addresses this with a shared-interface
core-replacement experiment. Holding the visual encoder, the FiLM mechanism, the action decoder and the action representation fixed, we replaced the circuit with six alternatives: the full circuit, an
MLP, an LSTM, a transformer and a reservoir, and additionally with a reservoir read out by a trained linear decoder.

The high-dimensional motor states of these variants were projected onto a three-dimensional manifold fitted to the clean trajectories of each model (Fig.~\ref{fig3}a). Under increasing noise the full circuit's noisy trajectories stayed close to its clean manifold, while the replacement cores drifted away and dispersed. The speed profiles on the manifold (Fig.~\ref{fig3}b) show the same asymmetry: the
full circuit's speed band around the clean profile is 0.03, compared with 0.62 (MLP), 0.28 (LSTM), 0.35 (transformer), 0.59 (reservoir) and 0.58 (reservoir read-out). Five trajectory-shape and dynamics metrics (Fig.~\ref{fig3}c) reproduce this ordering across the noise sweep, with effect sizes that grow with noise. The advantage is thus not a property of the interface, of parameter count or of recurrence in general; it is carried by the specific circuit.

\subsection{Robustness and ablation effects transfer to a real robot}

\begin{figure}[htbp]
\centering
\includegraphics[width=\textwidth]{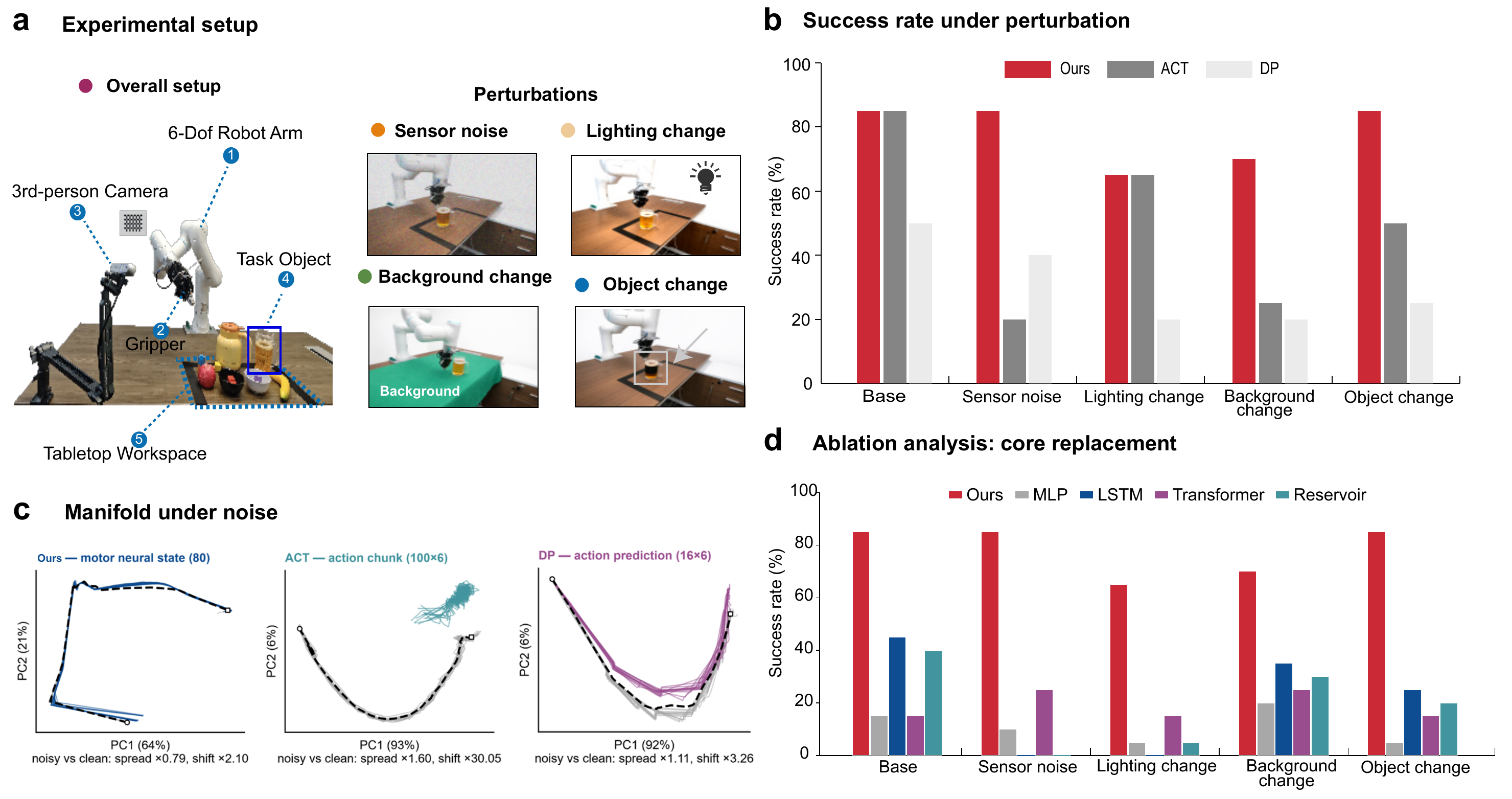}
\caption{\textbf{Robustness, deviation dynamics and ablation on a real robot.} \textbf{a}, Experimental setup and the four perturbations (sensor noise, lighting change, background change and object change).
\textbf{b}, Success rate of the core, ACT and DP under the nominal condition and under each perturbation (Ours red, ACT grey, DP light grey). \textbf{c}, Manifold of each policy's own representation (Ours motor neural state, ACT action chunk, DP action prediction) projected on the first two principal components fitted to that model's clean repeats; annotations give the ratio of noisy to clean trial-to-trial spread and the center shift. \textbf{d}, Ablation analysis by core replacement: success rate of the MLP, LSTM, transformer and reservoir replacements together with the full core (Ours red) under the nominal condition and each perturbation.}
\label{fig4}
\end{figure}

The final question is whether the picture survives on physical hardware. We deployed the core and the baselines on a real robot performing a cup-push task from a fixed initial pose (Fig.~\ref{fig4}a). In the nominal condition the core succeeded in 18/20 trials, ACT in 16/20 and DP in 14/20. Under the four perturbations the ordering became decisive (Fig.~\ref{fig4}b): the
core retained 15--16/20 whereas ACT fell to 8--14/20 and DP to 0--12/20, with the largest separations under illumination change (15/8/0) and object displacement (15/9/6). NCP was excluded because its nominal real-robot success was already low (13/20).

The commanded action shows the same asymmetry (Fig.~\ref{fig4}c): the noisy trials of ACT and DP leave the clean manifold (spread/shift
0.79/2.10, 1.60/30.05 and 1.11/3.26). The same ranking appears when the core is replaced on the robot (Fig.~\ref{fig4}d): the full circuit is the only variant above 50\% under every perturbation, whereas MLP, LSTM, transformer and reservoir replacements stay at or below 45\% and lose most or all perturbed trials.

\section{Discussion}

Our results support three statements. The first is that a biophysically detailed biological sensorimotor circuit can act as a task-agnostic dynamical core: one circuit, constrained by biological data and held its weights fixed, supported multiple simulated manipulation tasks through thin task-specific adapters. The second is attribution. Because the interface, data, optimizer and evaluation were held fixed, and because replacing the core with MLP, LSTM, transformer or reservoir alternatives removed the advantage, the effect localizes to the circuit rather than to the adapter, to parameter count or to recurrence as such. The third concerns real-world transferability: robustness and ablation effects of the core can transfer well to a real robot.

These findings connect to two literatures. In neuroscience, the idea that nervous systems exploit low-dimensional population dynamics to generate behaviour is well established for reaching and other motor tasks \citep{churchland2012neural,cunningham2014dimensionality,gallego2017neural,vyas2020computation}, and recurrent networks trained on tasks reproduce many of those geometric signatures \citep{sussillo2013opening,maheswaranathan2019universality}. Our results add a complementary observation: when the recurrent substrate is a real connectome with biologically constrained cell models, the resulting dynamics are not only interpretable but also transferable to a body that the circuit never evolved for. In machine learning, the same result speaks to the value of structured priors \citep{jaeger2004harnessing,maass2002real,lukosevicius2009reservoir} and specifically to the observation that recurrent processing confers robustness that feed-forward architectures lack
\citep{kietzmann2019recurrence}. A fixed-parameter biological circuit is a strong version of this prior: it is not a regularizer added to a task network but the generator of the task's sensorimotor trajectories.

Several limitations bound these claims. The number of real-robot tasks is small: we report one manipulation task with a fixed initial pose, and although the perturbations are physically realistic, they do not exhaust the space of distribution shifts. Second, the biological circuit is reused rather than created here, and its behavioural repertoire is that of a small sensorimotor system; how far the same principle extends to tasks requiring long-horizon planning or dexterous manipulation is an open question.

The practical implication is nevertheless concrete. If a biophysically detailed circuit supplies the dynamical core, the fragile part of a robot-learning pipeline shrinks to the adapters that map observations into the core and its state into actions, and robustness and interpretability follow from the same experimentally constrained object.


\section{Methods}

\subsection{Biological core}
The core is the brain model of BAAIWorm \citep{zhao2024baaiworm}: 136 neurons (15 sensory, 41 interneurons and 80 motor neurons) with realistic morphology, multicompartment dynamics (Hodgkin--Huxley
somas, passive neurites, fourteen ion-channel families including voltage-gated calcium channels and calcium-regulated potassium channels) and the worm's connectome, comprising 1{,}901 inter-neuron
connections (1{,}115 chemical synapses and 786 gap junctions) plus 15 sensory-input connections. The core is simulated in continuous time; its synaptic weights are fixed in every experiment reported here, whereas its membrane voltages evolve freely.

\subsection{Interface and training}
Four consecutive RGB frames are encoded by a small convolutional network into a 15-dimensional sensory drive. The drive is scaled into clamp currents and injected into the sensory neurons; the simulation is
advanced between control steps and the 80 motor-neuron voltages form the motor neural state. A FiLM mechanism conditions this state on the visual feature, and a multi-layer perceptron decodes an eight-step
action chunk of which four steps are executed before the next observation is processed (predict~8, execute~4, re-observe and replan). Only the encoder, FiLM parameters and decoder are trained,
by behaviour cloning from demonstrations; the core receives no gradient updates.

\subsection{Baseline policies}
DP \citep{chi2023diffusion} and ACT \citep{zhao2023act} were trained with their original recipes on the same demonstrations and the same action representation. NCP \citep{lechner2020ncp} was trained as a native system with its own interface and is therefore reported as a system-level baseline rather than as a core replacement. For the shared-interface core replacement, the encoder, FiLM mechanism, decoder, optimizer, data and checkpoint-selection rule were held fixed while the core was replaced by: an MLP, an LSTM, a transformer, a reservoir, and a reservoir with a trained linear read-out.

\subsection{Simulated tasks and perturbations}
MetaWorld \citep{yu2020meta} provided coffee-push, hammer, door-open and assembly. Three perturbations were applied: pixel-level Gaussian noise at a single intensity ($\sigma = 0.10$), illumination change and background change. Success rates were measured with 50 evaluations. The core-replacement used the same encoder, FiLM mechanism and decoder, and the manifold and speed analyses were computed on the motor states of the same rollouts.

\subsection{Manifold, speed and recovery analyses}
Motor states from clean and perturbed rollouts were standardized and projected onto the three leading principal components fitted to the clean trajectories of the corresponding model. We quantified the
displacement of noisy trajectories from the clean manifold, the band width of the speed profile around the clean profile, and five trajectory metrics (temporal jitter, normalized jerk, path curvature,
velocity dissimilarity and tangling), each normalized to its clean reference.

\subsection{Real-robot protocol and statistics}
A six-degree-of-freedom arm with a parallel gripper performed a cup-push task from a fixed initial pose, driven by the policies' own outputs at the control rate. Each policy was evaluated in 20 trials under the nominal condition and under four perturbations: additive image noise ($\sigma = 30$), turning off three light sources, replacing the background with a green cloth, and displacing the task objects. Success was judged by an on-site annotator against a fixed criterion.


\bibliographystyle{unsrtnat}
\bibliography{references}

@article{white1986connectome,
  author  = {White, J. G. and Southgate, E. and Thomson, J. N. and Brenner, S.},
  title   = {The structure of the nervous system of the nematode \textit{Caenorhabditis elegans}},
  journal = {Philosophical Transactions of the Royal Society of London. B},
  volume  = {314},
  number  = {1165},
  pages   = {1--340},
  year    = {1986},
  doi     = {10.1098/rstb.1986.0056}
}

@article{varshney2011structural,
  author  = {Varshney, L. R. and Chen, B. L. and Paniagua, E. and Hall, D. H. and Chklovskii, D. B.},
  title   = {Structural properties of the \textit{Caenorhabditis elegans} neuronal network},
  journal = {PLoS Computational Biology},
  volume  = {7},
  number  = {2},
  pages   = {e1001066},
  year    = {2011},
  doi     = {10.1371/journal.pcbi.1001066}
}

@article{towlson2013rich,
  author  = {Towlson, E. K. and V{\'e}rtes, P. E. and Ahnert, S. E. and Schafer, W. R. and Bullmore, E. T.},
  title   = {The rich club of the \textit{C.~elegans} neuronal connectome},
  journal = {Journal of Neuroscience},
  volume  = {33},
  number  = {15},
  pages   = {6380--6387},
  year    = {2013},
  doi     = {10.1523/JNEUROSCI.3784-12.2013}
}

@article{cook2019whole,
  author  = {Cook, S. J. and Jarrell, T. A. and Brittin, C. A. and Wang, Y. and Bloniarz, A. E. and Yakovlev, M. A. and Nguyen, K. C. Q. and Tang, L. T.-H. and Bayer, E. A. and Duerr, J. S. and B{\"u}low, H. E. and Hobert, O. and Hall, D. H. and Emmons, S. W.},
  title   = {Whole-animal connectomes of both \textit{Caenorhabditis elegans} sexes},
  journal = {Nature},
  volume  = {571},
  pages   = {63--71},
  year    = {2019},
  doi     = {10.1038/s41586-019-1352-7}
}

@article{zhao2024baaiworm,
  author  = {Zhao, M. and Wang, N. and Jiang, X. and Ma, X. and Ma, H. and He, G. and Du, K. and Ma, L. and Huang, T.},
  title   = {An integrative data-driven model simulating \textit{C.~elegans} brain, body and environment interactions},
  journal = {Nature Computational Science},
  volume  = {4},
  pages   = {978--990},
  year    = {2024},
  doi     = {10.1038/s43588-024-00738-w}
}

@misc{chi2023diffusion,
  author    = {Chi, C. and Feng, S. and Du, Y. and Xu, Z. and Cousineau, E. and Burchfiel, B. and Song, S.},
  title     = {Diffusion policy: visuomotor policy learning via action diffusion},
  year      = {2023},
  eprint    = {2303.04137},
  note      = {In Robotics: Science and Systems XIX. \url{https://doi.org/10.15607/RSS.2023.XIX.026}}
}

@misc{zhao2023act,
  author    = {Zhao, T. Z. and Kumar, V. and Levine, S. and Finn, C.},
  title     = {Learning fine-grained bimanual manipulation with low-cost hardware},
  year      = {2023},
  eprint    = {2304.13705},
  note      = {In Robotics: Science and Systems XIX. \url{https://doi.org/10.15607/RSS.2023.XIX.016}}
}

@article{lechner2020ncp,
  author  = {Lechner, M. and Hasani, R. and Amini, A. and Henzinger, T. A. and Rus, D. and Grosu, R.},
  title   = {Neural circuit policies enabling auditable autonomy},
  journal = {Nature Machine Intelligence},
  volume  = {2},
  pages   = {642--652},
  year    = {2020},
  doi     = {10.1038/s42256-020-00237-3}
}

@misc{yu2020meta,
  author    = {Yu, T. and Quillen, D. and He, Z. and Julian, R. and Hausman, K. and Finn, C. and Levine, S.},
  title     = {Meta-World: a benchmark and evaluation for multi-task and meta reinforcement learning},
  year      = {2019},
  eprint    = {1910.10897},
  note      = {Preprint at \url{https://arxiv.org/abs/1910.10897}}
}

@misc{tobin2017domain,
  author    = {Tobin, J. and Fong, R. and Ray, A. and Schneider, J. and Zaremba, W. and Abbeel, P.},
  title     = {Domain randomization for transferring deep neural networks from simulation to the real world},
  year      = {2017},
  eprint    = {1703.06907},
  note      = {In IEEE/RSJ International Conference on Intelligent Robots and Systems (IROS), 23--30. \url{https://doi.org/10.1109/IROS.2017.8202133}}
}

@misc{hendrycks2019benchmarking,
  author    = {Hendrycks, D. and Dietterich, T.},
  title     = {Benchmarking neural network robustness to common corruptions and perturbations},
  year      = {2019},
  eprint    = {1903.12261},
  note      = {Preprint at \url{https://arxiv.org/abs/1903.12261}}
}

@article{churchland2012neural,
  author  = {Churchland, M. M. and Cunningham, J. P. and Kaufman, M. T. and Foster, J. D. and Nuyujukian, P. and Ryu, S. I. and Shenoy, K. V.},
  title   = {Neural population dynamics during reaching},
  journal = {Nature},
  volume  = {487},
  pages   = {51--56},
  year    = {2012},
  doi     = {10.1038/nature11129}
}

@article{cunningham2014dimensionality,
  author  = {Cunningham, J. P. and Yu, B. M.},
  title   = {Dimensionality reduction for large-scale neural recordings},
  journal = {Nature Neuroscience},
  volume  = {17},
  pages   = {1500--1509},
  year    = {2014},
  doi     = {10.1038/nn.3776}
}

@article{gallego2017neural,
  author  = {Gallego, J. A. and Perich, M. G. and Miller, L. E. and Solla, S. A.},
  title   = {Neural manifolds for the control of movement},
  journal = {Neuron},
  volume  = {94},
  pages   = {978--984},
  year    = {2017},
  doi     = {10.1016/j.neuron.2017.05.025}
}

@article{vyas2020computation,
  author  = {Vyas, S. and Golub, M. D. and Sussillo, D. and Shenoy, K. V.},
  title   = {Computation through neural population dynamics},
  journal = {Annual Review of Neuroscience},
  volume  = {43},
  pages   = {249--275},
  year    = {2020},
  doi     = {10.1146/annurev-neuro-092619-094115}
}

@article{sussillo2013opening,
  author  = {Sussillo, D. and Barak, O.},
  title   = {Opening the black box: low-dimensional dynamics in high-dimensional recurrent neural networks},
  journal = {Neural Computation},
  volume  = {25},
  pages   = {626--649},
  year    = {2013},
  doi     = {10.1162/NECO_a_00409}
}

@misc{maheswaranathan2019universality,
  author    = {Maheswaranathan, N. and Williams, A. H. and Golub, M. D. and Ganguli, S. and Sussillo, D.},
  title     = {Universality and individuality in neural dynamics across large populations of recurrent networks},
  year      = {2019},
  eprint    = {1907.08549},
  note      = {Preprint at \url{https://arxiv.org/abs/1907.08549}}
}

@article{jaeger2004harnessing,
  author  = {Jaeger, H. and Haas, H.},
  title   = {Harnessing nonlinearity: predicting chaotic systems and saving energy in wireless communication},
  journal = {Science},
  volume  = {304},
  pages   = {78--80},
  year    = {2004},
  doi     = {10.1126/science.1091277}
}

@article{maass2002real,
  author  = {Maass, W. and Natschl{\"a}ger, T. and Markram, H.},
  title   = {Real-time computing without stable states: a new framework for neural computation based on perturbations},
  journal = {Neural Computation},
  volume  = {14},
  pages   = {2531--2560},
  year    = {2002},
  doi     = {10.1162/089976602760407955}
}

@article{lukosevicius2009reservoir,
  author  = {Luko{\v{s}}evi{\v{c}}ius, M. and Jaeger, H.},
  title   = {Reservoir computing approaches to recurrent neural network training},
  journal = {Computer Science Review},
  volume  = {3},
  pages   = {127--149},
  year    = {2009},
  doi     = {10.1016/j.cosrev.2009.03.005}
}

@article{kietzmann2019recurrence,
  author  = {Kietzmann, T. C. and Spoerer, C. J. and S{\"o}rensen, L. K. A. and Cichy, R. M. and Haan, R. and Kriegeskorte, N.},
  title   = {Recurrence is required to capture the representational dynamics of the human visual system},
  journal = {Proceedings of the National Academy of Sciences},
  volume  = {116},
  pages   = {21854--21863},
  year    = {2019},
  doi     = {10.1073/pnas.1905544116}
}

\end{document}